\documentclass{article}
\usepackage{spconf,amsmath,graphicx}

\usepackage{amsmath,amsfonts,bm, bbm}

\def\eqref#1{equation~\ref{#1}}

\def\1{\bm{1}}

\def\rvb{{\mathbf{b}}}

\def\rmV{{\mathbf{V}}}
\def\rmW{{\mathbf{W}}}

\def\vg{{\bm{g}}}
\def\vh{{\bm{h}}}

\def\vx{{\bm{x}}}
\def\vy{{\bm{y}}}

\def\mI{{\bm{I}}}

\DeclareMathAlphabet{\mathsfit}{\encodingdefault}{\sfdefault}{m}{sl}
\SetMathAlphabet{\mathsfit}{bold}{\encodingdefault}{\sfdefault}{bx}{n}

\usepackage{mdframed}

\newtheorem{thm}{Theorem}

\usepackage{accents}
\newlength{\dhatheight}

\usepackage{cite}
\usepackage{lipsum}
\usepackage{adjustbox}    
\usepackage[dvipsnames]{xcolor}
\usepackage[utf8]{inputenc} 
\usepackage[T1]{fontenc}    
\usepackage{hyperref}       
\usepackage{url}            
\usepackage{booktabs}       
\usepackage{amsfonts}       
\usepackage{nicefrac}       
\usepackage[position=top]{subfig}
\usepackage{caption}
\usepackage{tikz}
\usepackage{enumitem}
\usepackage{soul}
\usepackage{tabularx}
\usetikzlibrary{er,positioning}

\title{NFT-K: Non-Fungible Tangent Kernels}
\name{Sina Alemohammad, Hossein Babaei, CJ Barberan, Naiming Liu,}{Lorenzo Luzi, Blake Mason, Richard G.\ Baraniuk}
\address{Rice University}
\usepackage{siunitx}
\usepackage{tikz}
\usetikzlibrary{calc}
\usepackage{pgfplots}
\usepackage{pgfplotstable}
\usepackage{readarray}
\usepackage{booktabs}
\usepackage{array}
\usepackage{multirow}
\usepackage{multicol}
\usetikzlibrary{positioning, fit, shapes.geometric,arrows}
\pgfplotsset{compat=1.15}
\usetikzlibrary{datavisualization}
\usetikzlibrary{datavisualization.formats.functions}
\usepgfplotslibrary{fillbetween}

\newcommand\blfootnote[1]{%
  \begingroup
  \renewcommand\thefootnote{}\footnote{#1}%
  \addtocounter{footnote}{-1}%
  \endgroup
}

\begin{document}
%
\maketitle
\begin{abstract}
Deep neural networks have become essential for numerous applications due to their strong empirical performance such as vision, RL, and classification. Unfortunately, these networks are quite difficult to interpret, and this limits their applicability in settings where interpretability is important for safety, such as medical imaging. One type of deep neural network is neural tangent kernel that is similar to a kernel machine that provides some aspect of interpretability. To further contribute interpretability with respect to classification and the layers, we develop a new network as a combination of multiple neural tangent kernels, one to model each layer of the deep neural network individually as opposed to past work which attempts to represent the entire network via a single neural tangent kernel. We demonstrate the interpretability of this model on two datasets, showing that the multiple kernels model elucidates the interplay between the layers and predictions.  \blfootnote{This work was supported by NSF grants CCF-1911094, IIS-1838177, and IIS-1730574; ONR grants N00014-18-12571, N00014-20-1-2787, and N00014-20-1-2534; AFOSR grant FA9550-18-1-0478; and a Vannevar Bush Faculty Fellowship, ONR grant N00014-18-1-2047. \nonumber
}

\end{abstract}
\section{Introduction}
\label{sec:intro}

Deep neural networks (DNNs) have become the main parametric models used to solve machine learning problems as they provide state-of-the-art empirical performance, e.g., see~\cite{brock2018large,karras2020analyzing,szegedy2016rethinking,simonyan2014very,he2016deep,silver2016mastering}. 
DNNs can be used for a host of applications, including autonomous driving~\cite{uccar2017object}, medical imaging~\cite{ronneberger2015u}, medical records~\cite{shickel2017deep}, and cyber security~\cite{dixit2021deep}.
These applications all benefit from DNNs but are also high risk situations.

Despite the extraordinary performance of DNNs, it is still not clear how to best interpret their predictions and inferences. We direct the reader to this summary of recent efforts on interpretations~\cite{gilpin2018explaining}. 
Indeed there are some aspects within DNN interpretability that need more understanding or are poorly explained by prior work. A core challenge is to understand which of the layers are responsible for the which aspects of the decision-making process for each class.

Towards a deeper understanding of the functionality of neural networks, recent work has demonstrated that 
some DNNs correspond to kernel machines~\cite{friedman2017elements}. Kernel machines have been studied extensively and can provide extra interpretability depending on the structure of the kernel.
Similar to their kernel counterparts, DNN architectures typically use a large number of parameters which can be orders of magnitude more than the training samples. Overparameterized DNNs generalize unseen data very well despite over-fitting training data, which is a property of interpolating methods such as kernel machines \cite{belkin2019reconciling, belkin2018understand}. 
The exact relationship between DNNs and kernel methods have been revealed by analysing the {\em infinite-width} DNNs, when the number of units (neurons) in the hidden layers goes to infinity. It has been shown that infinite-width DNNs trained with gradient descent converges to a kernel ridge regression predictor with respect to a kernel known as the {\em neural tangent kernel} (NTK) \cite{jacot2018neural}. In recent years NTKs have been developed for a variety of architectures \cite{arora2019exact,novak2018bayesian,du2019graph,huang2020deep,hron2020infinite,alemohammad2020recurrent,alemohammad2020scalable, yang2020tensor}.




Though they live in potentially infinite dimensional spaces, kernel predictors can be reduced to linear classifiers in the eigenspace of the grammian of the input data. This is the famed ``kernel trick'' in machine learning and allows improved computational efficiency and interpretability of results. 
NTKs likewise use a gram matrix, it is easier to interpret the relationship between the input and the output of the DNN. However, it is not straightforward to use NTK to interpret which \textit{layers} the network uses to classify a certain class via the gram matrix. The relationship between class and layer is not well understood but can be useful for identifying when misclassifications occur or if depth should be increased or decreased in a neural network. 
We begin to understand the relationship between classes and layers by modelling it using multiple NTKs, one for each layer.

\subsection{Contributions}

\begin{figure}[t]
    \centering
    \includegraphics[width=\linewidth]{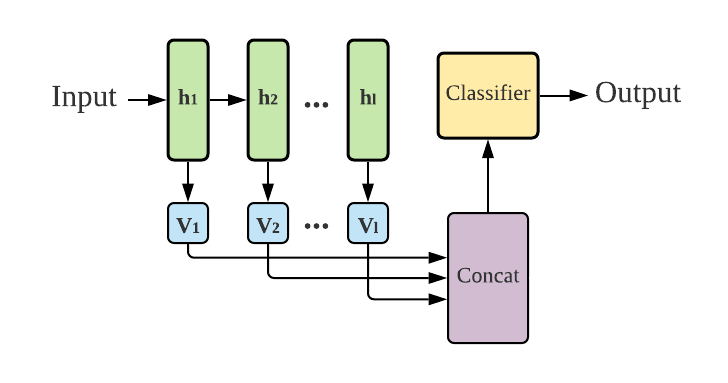}
    \caption{Diagram of the NFT-K MLP network. Each inner product of the kernel with the layer is sent as the input to the classifier. The weights of the classifier can interpret which layers are aiding in the decision-making process.}
    \label{fig:nft}
\end{figure}

In order to capture the relationship between layers and classes, we combine NTK with inspiration from another network~\cite{lin2017feature}. Our architecture admits multiple-kernel learning in the infinite-width limit, where each kernel corresponds to the NTK of each layer. We use the layers of multi-layer perceptrons (MLPs) and call our model the non-fungible tangent kernel MLP (NFT-K MLP) for our proposed architecture. This stems from our model learning to select layers of a DNN which are crucial for classification of certain classes (Figure~\ref{fig:nft}). As the layers selected by the model are the most important, they are not fungible with the layers that were not selected, and this reduces the overall complexity of the model while ensuring strong empirical performance. 

To validate our model, we conduct experiments on the MNIST dataset and the large movie review dataset to provide interpretable results into how our network provides insight into which layers are important for classification. We see that NFT-K MLPs focus on similar layers for classes which visually appear similar, even without the use of convolutional layers. Moreover, we can see which layers are the most important for classification; some layers are not at all important and are dropped from the model. Overall, NFT-K MLPs are able to model the relationship between classes and layers and provide insightful information.

\section{Background}
Given an input $\vx \in \mathbb{R}^{m}$, an MLP with $L$ hidden layers and $n$ neurons performs the following recursive relations
\begin{align}
    \vg^{(1)}(\vx) &= \rmW^{(1)}\vx + \rvb^{\left(1\right)}, \label{r1}\\
    \vh^{(\ell)}(\vx) &= \phi(\vg^{(\ell)}(\vx)), \label{r2}\\
    \vg^{(\ell)}(\vx) &= \rmW^{(\ell)}\vh^{(\ell-1)}(\vx) + \rvb^{\left(\ell\right)}, \quad 2\leq \ell \leq L \label{r3},
\end{align}
where $\rmW^{(1)} \in \mathbb{R}^{m\times n}$, $\rmW^{(\ell)} \in \mathbb{R}^{n\times n}$, and $\rvb^{\left(\ell\right)} \in \mathbb{R}^{n}$.  $\phi(\cdot): \mathbb{R} \rightarrow \mathbb{R}$ represents the activation function that acts coordinate wise on a vector.

The output of an MLP is obtained by a linear transformation of the last layer 
\begin{align}
     f_{\theta}^{(L)}(\vx)&= \rmV^{(L)}\vh^{(L)}(\vx) \in \mathbb{R}^{d}, \label{out}
\end{align}
where $\theta$ is the set of all parameters in the networks. 
Let $\mathcal{X}$  be the set of training inputs. Let and $\mathcal{Y} \in \mathbb{R}^{| \mathcal{X}|d}$ be the concatenation of all labels. 
Let $l(\widehat{y},y): \mathbb{R}^d\times \mathbb{R}^d \rightarrow \mathbb{R}^+$ be the MSE loss function, and $\mathcal{L}(\theta) = \frac{1}{{|\mathcal{X}|} } \sum_{(\vx,\vy) \in \mathcal{X}\times \mathcal{Y}} l(f_{\theta}(\vx), \vy) $ be the the empirical loss. Let $\theta_0$ be the networks parameters at initialization. 
An MLP trained with infinitesimal step size for Gaussian distributed $\theta_0$ using the following loss
\begin{align}
    \min_{\theta} \Big(\mathcal{L}(\theta) + \lambda \| \theta - \theta_0  \|_2^2 \Big) 
\end{align}
is equivalent to a kernel ridge regression predictor with ridge hyperparameter $\lambda$ and a deterministic kernel known as its neural tangent kernel (NTK). The NTK is defined as 
\begin{align}
    \Theta(\vx,\vx') = \Theta^{(L)}(\vx,\vx') \mI_d
    = \nabla_{\theta} f^{(L)}_\theta(\vx)^\top  \nabla_{\theta} f^{(L)}_\theta(\vx')
    \label{fml80}
\end{align}
where $\Theta^{(L)}(\vx,\vx')  \in \mathbb{R}$ is the NTK of a MLP with depth $L$ that depends on the network's activation function, depth and initialization hyperparameters~\cite{jacot2018neural}. We define $\Theta^{(L)}(\vx,\mathcal X)$ as the vector in $\mathbb R^{|\mathcal{X}|}$ whose entries are $\Theta^{(L)}(\vx,\vx')$ for $\vx' \in \mathcal X$; similarly, $\Theta^{(L)}(\mathcal X,\mathcal X') \in \mathbb R^{|\mathcal X|\times |\mathcal X'|}$.

\label{sec:background}
\section{Non-Fungible Tangent Kernels}

In this section, we propose our NFT-K architecture. Consider the recursive relations from Equations (\ref{r1}), (\ref{r2}), and (\ref{r3}) that form the hidden states $\vh^{(\ell)}(\vx)$. Instead of obtaining the output by a linear transformation from the last layer as shown in Equation (\ref{out}), we obtain the output $f_{P}(\vx)$:
\begin{align}
     f^{(\ell)}(\vx) &= \rmV^{(\ell)}\vh^{(\ell)}(\vx), \\
     f(\vx) &=  \sum_{\ell = 1}^{L} \omega_{\ell} \odot f_{P}^{(\ell)}(\vx) \quad \omega_{\ell} \in \mathbb{R}^d,
\end{align} 
where $\odot$ is the entry-wise product. Note that $\rmV^{(\ell)}$ for $\ell = 1, \dots, L-1$ are not present in the MLP architecture.
Figure~\ref{fig:nft} displays a network diagram for visualizing of this architecture. 

We are interested in the following optimization problem 
\begin{align}
    \min_{\Omega} \min_{\theta} \Big(\mathcal{L}(\theta, \Omega) + \lambda \| \theta - \theta_0  \|_2^2 \Big) \quad s.t \quad  \Omega \in \mathcal{S} 
    \label{fml9}
\end{align}
where 
\begin{align}
    \theta &= \mathrm{Vect}[\{\rmW^{(\ell)},\rmV^{(\ell)},\rvb^{(\ell)}\}_{\ell = 1}^{L}], \\
    \Omega &=  [ \omega_{\ell}, \dots, \omega_{L} ] \in \mathbb{R}^{d \times L}, \\
    \mathcal{S} &= \left\{ \Omega \in \mathbb{R}^{d\times L} | \Omega > 0  \right\}.
\end{align}
To solve (\ref{fml9}), we use alternating minimization by first solving over $\theta$ for a fixed $\Omega$ and then solving over $\Omega$ for a fixed $\theta$. More precisely, we perform gradient descent on $\theta$ for fixed $\Omega$:
\begin{align}
    \theta_{t+1} = \theta_{t} - \eta_{\theta}\nabla_{\theta_t}  \mathcal{L}(\theta_t, \Omega) - \eta_{\theta}\lambda(\theta_t - \theta_0) \label{fml3}   
\end{align}
until convergence. In practice, we use stochastic gradient descent with mini-batches. The following theorem characterizes the output of the network after each update performed in Equation~\ref{fml3}.
\begin{thm} \label{fml2}
After convergence of gradient descent in (\ref{fml3}) over $\theta$ for a fixed $\Omega$ with infinitesimal step size $\eta_{\theta} \rightarrow 0 $, the output of the neural network will be
\begin{align}
     f(\vx) = \Theta_{\Omega}(\vx,\mathcal{X})\left( \Theta_{\Omega}(\mathcal{X},\mathcal{X}) + \lambda\mI_{d|\mathcal{X}|}   \right)\mathcal{Y}, \label{fml5}
\end{align}
where 
\begin{align}
    \Theta_{\Omega}(\vx,\vx') = \sum_{\ell = 1}^{L} \Theta^{(\ell)}(\vx,\vx') \mathrm{diag}(\omega_{\ell}) \label{lolo}
\end{align}
and $\Theta^{(\ell)}(\vx,\vx')$ is the NTK of an MLP with depth $\ell$ as defined in Equations \ref{r1} to \ref{fml80}.
\end{thm}
\emph{Proof of Theorem~\ref{fml2}.} For a fixed $\Omega$, the output of the network is just a linear combination of the output at each layers. Since we are using gradient descent with infinitesimal step size the out put will be kernel ridge regression with respect to the NTK of network and the NTK can be easily obtained from the results of \cite{yang2020tensor}. $\square$

In particular this result highlights that in the infinite width limit, solving over $\theta$ has a closed form solution that reduces to kernel ridge regression. The corresponding kernel is a linear combination of the NTKs of each layer of the network, as shown in Equation \ref{lolo}.  

In the second step of our alternating minimization procedure, optimizing over $\Omega \in \mathcal{S}$ is a very well known problem known as multiple kernel learning \cite{zhang2016fast}. Hence, optimizing the network loss in \ref{fml9} using alternating minimization results in multiple-kernel learning in the infinite-width limit and the correspondingly learned $\omega_{\ell}$ informs us of each layer's importance in the given classification task. 

To solve over $\Omega$, we use gradient projection:
\begin{align}
    \Omega_{t+1} = \mathcal{P}_{\mathcal{S}}\left(\Omega_{t} - \eta_{\Omega} \nabla_{\Omega_{t}}\mathcal{L}(\theta, \Omega)\right). \label{fml4}
\end{align}


\begin{table}[t]
    \centering
    \caption{ \small
    Test accuracy among different datasets showing that MLPs and NFT-K MLP perform similarly. In the MNIST classification task, the 10 layer NFT-K MLP outperforms all other networks. In the large movie review classification problem, MLPs and NFT-K MLP perform within 0.42\% of one another.}
        {\small 
    \begin{tabular}{lcc} \hline
        \toprule
         Network &   Dataset  & Test Acc \\ 
         \midrule
         10 layer MLP & MNIST & 96.89\% \\
         10 layer NFT-K MLP & MNIST & \textbf{97.25}\% \\ 
         20 layer MLP & MNIST & 97.22\% \\ 
         20 layer NFT-K MLP & MNIST & 95.23\% \\
         \hline 
         
         5 layer MLP & Large Movie Review & \textbf{86.73}\% \\
         5 layer NFT-K MLP & Large Movie Review & 86.4\% \\
         10 layer MLP & Large Movie Review & 86.31\% \\
         10 layer NFT-K & Large Movie Review & 86.7\% \\

         \bottomrule

    \end{tabular}
    }
    \label{tab:results}

\end{table}

\begin{figure*}[t]

\pgfplotsset{
        heataxis/.style={
          width=5.5cm,
            y dir = reverse,
            colormap/blackwhite
          }
        }

\def\figwidth{0.28}
\centering

\begin{minipage}{\figwidth\textwidth}
    \centering
    \subfloat{

    \begin{tikzpicture}[trim axis left, trim axis right]
        \begin{axis}[heataxis, 
            title={Class importance (10 layers)}, xlabel=Class, ylabel=Class,
            xmin=-0.5,
            xmax=9.5,
            ymin =-0.5,
            ymax=9.5]
            
            \addplot [matrix plot*,point meta=explicit] file [meta=index 2]  {csv/cc10.csv};
        \end{axis}
    \end{tikzpicture}
    } 
\end{minipage}
\begin{minipage}{\figwidth\textwidth}
    \centering
    \subfloat{
    
    \begin{tikzpicture}[trim axis left, trim axis right]
        \begin{axis}[heataxis, 
            title={Layer importance (10 layers)}, xlabel=Layer, ylabel=Layer,
            xmin=0.5,
            xmax=10.5,
            ymin =0.5,
            ymax=10.5]
            
            \addplot [matrix plot*,point meta=explicit] file [meta=index 2]  {csv/ll10.csv};
        \end{axis}
    \end{tikzpicture}
    } 
\end{minipage}
\begin{minipage}{\figwidth\textwidth}
    \centering
    \subfloat{\
    
    \begin{tikzpicture}[trim axis left, trim axis right]
        \begin{axis}[heataxis, 
            title={Class-layer importance (10 layers)}, xlabel=Layer, ylabel=Class,
            colorbar,
            xmin=0.5,
            xmax=10.5,
            ymin =-0.5,
            ymax=9.5]
            
            \addplot [matrix plot*,point meta=explicit] file [meta=index 2]  {csv/lc10.csv};
        \end{axis}
    \end{tikzpicture}
    }
\end{minipage}
\\
\begin{minipage}{\figwidth\textwidth}
    \centering
    \subfloat{

    \begin{tikzpicture}[trim axis left, trim axis right]
        \begin{axis}[heataxis, 
            title={Class importance (20 layers)}, xlabel=Class, ylabel=Class,
            xmin=-0.5,
            xmax=9.5,
            ymin =-0.5,
            ymax=9.5]
            
            \addplot [matrix plot*,point meta=explicit] file [meta=index 2]  {csv/cc20.csv};
        \end{axis}
    \end{tikzpicture}
    
    } 
\end{minipage}
\begin{minipage}{\figwidth\textwidth}
    \centering
    \subfloat{
    
    \begin{tikzpicture}[trim axis left, trim axis right]
        \begin{axis}[heataxis, 
            title={Layer importance (20 layers)}, xlabel=Layer, ylabel=Layer,
            xmin=0.5,
            xmax=20.5,
            ymin =0.5,
            ymax=20.5]
            
            \addplot [matrix plot*,point meta=explicit] file [meta=index 2]  {csv/ll20.csv};
        \end{axis}
    \end{tikzpicture}} 
\end{minipage}
\begin{minipage}{\figwidth\textwidth}
    \centering
    \subfloat{\
    
    \begin{tikzpicture}[trim axis left, trim axis right]
        \begin{axis}[heataxis, 
            title={Class-layer importance (20 layers)}, xlabel=Layer, ylabel=Class,
            colorbar,
            xmin=0.5,
            xmax=20.5,
            ymin =-0.5,
            ymax=9.5]
            
            \addplot [matrix plot*,point meta=explicit] file [meta=index 2]  {csv/lc20.csv};
        \end{axis}
    \end{tikzpicture}
    }
\end{minipage}
\caption{
We present heatmaps which can be used to interpret the class importance ($\omega$), layer importance, and class-layer importance. 
The heatmaps in the \textbf{top row} are constructed from a trained NFT-K MLP with 10 layers, while the heatmaps in the \textbf{bottom row} are constructed from a NFT-K MLP with 20 layers. We first define
$\Omega = \begin{bmatrix}\omega_1 & \hdots & \omega_L\end{bmatrix}$ and then normalize it by row to get $\Omega_n$. In this figure, the first column is $\Omega_n^\top\Omega_n$, the second column is $\Omega_n \Omega_n^\top$, and the third column is $\Omega_n$.
For the 10 layer case, 53.0\% of the elements of $\Omega$ are zero. For the 20 layer case, 45.5\% of the elements of $\Omega_n$ are zero.
}
\label{fig:mnist}
\end{figure*}
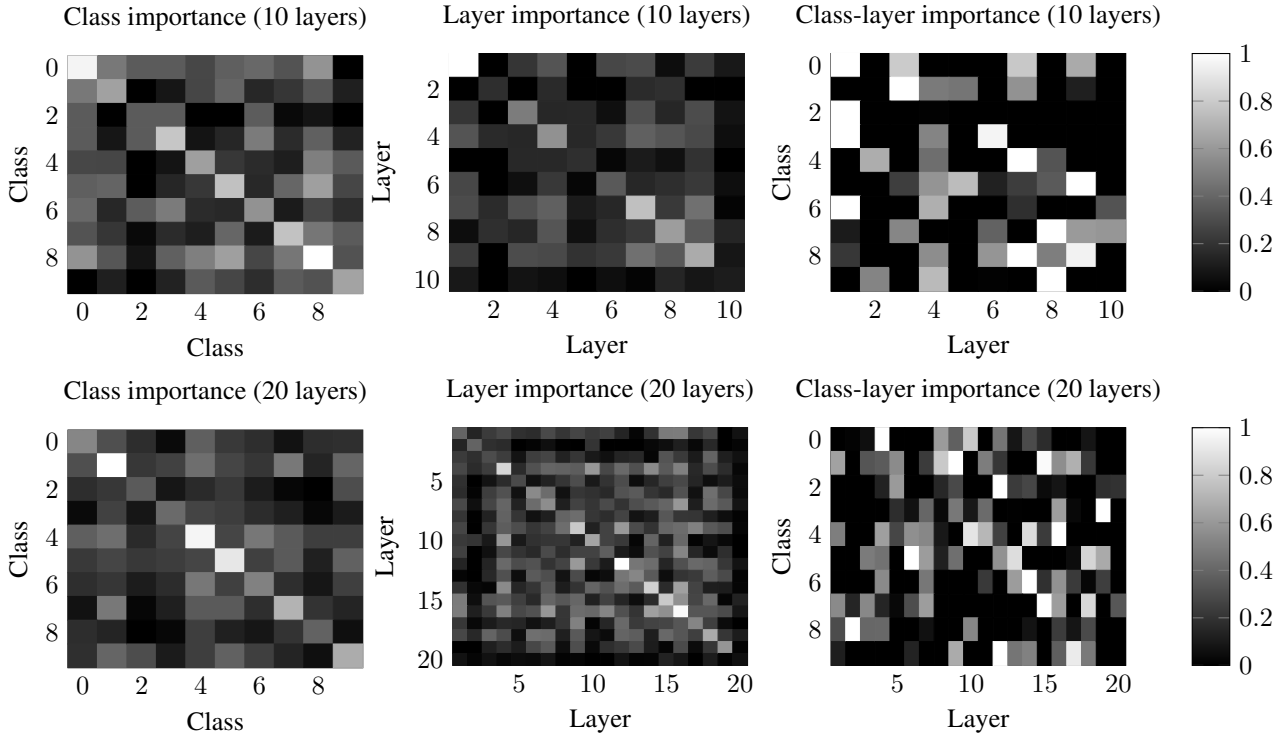

\section{Results}
\label{sec:result}
\subsection{Datasets}
The datasets used in the experiments are MNIST \cite{deng2012mnist} and Large Movie Review \cite{maas2011learning}. MNIST is composed 
of the handwritten digits from 0 to 9 and the task is to classify the 10 digits from the image. Large Movie Review dataset consists of movie reviews from IMDB and the task is to determine if the review's sentiment is positive or negative. For data preprocessing of this dataset, we use the encoder from the smallest GPT-2 model\cite{radford2019language} to extract word token embeddings. Then for each movie review, we normalize the embeddings based on the its length, and the resulting input embedding has a dimension of 768. 


\subsection{Hyperparameter Selection}

For MNIST in the NFT-K scenario we use 2 Stochastic Gradient Descent (SGD) \cite{le2011optimization} optimizers where one optimizer is for all the network parameters except the $\Omega$'s while the second optimizer is training the $\Omega$. we set $\eta_{\theta} = 0.02$ , while $\eta_{\Omega} = 0.05$, both with a weight decay of $2 \times 10^{-6}$. The number of neurons per layer are 256. The depths of the MLPs are 10 and 20. The mini-batch size is 50. The initialization strategy is Xavier normal initializer\cite{kumar2017weight}. For $\Omega$, they are projected to be positive.

For Large Movie Review, we use one SGD optimizer for all the network parameters. The learning rates is 0.1. The number of neurons per layer are 128. The depths of the MLPs are 5 and 10. The mini-batch size is 25. The experiment also uses Xavier normal initializer. For $\Omega$, they are projected to be positive.


\subsection{MNIST Experiment}
We compare an MLP network with an NFT-K MLP network with two different depths, as shown in Table~\ref{tab:results}. We show that the 10 layer NFT-K MLP outperforms all other networks that we trained on. Most of these networks performed similarly, even though the architectures are quite different. Therefore, in this experiment, NFT-K MLPs provide a way to increase interpretability without a cost to performance.

Our experiments show that, although classification was on par with the baseline, only one layer contributed to the classification of digit 2 in this experiment. We see from the top right panel of Figure \ref{fig:mnist} that the row for digit 2 has only a single non-zero entry. The non-zero entry is located at the first layer of the network, meaning that the proceeding layers do not contribute to the classification of 2 whatsoever.
In these plots, we calculate  $\Omega = \begin{bmatrix}\omega_1 & \hdots & \omega_L\end{bmatrix}$ and normalize its rows to get $\Omega_n$ because we want to compare how each class is being affected by each layer. 
We see that the sparsity of $\Omega_n$ is a useful property for determining which layers are important for classifying certain classes.

We can also see from the top right panel of Figure~\ref{fig:mnist} that the the rows for the digit 4 and 9 are similar. There's only one significant difference: layer 7 is critical for distinguishing between the two digits as it only appears when classifying the 4 digits. Moreover, we see that both layer 7 and layer 1 are crucial for classification (center top panel of Figure~\ref{fig:mnist}). It is obvious why layer 1 is important: it is the only layer that can classify the 2 digits and it plays a role in classification of 6 out of the 10 digits. Thus, we can use these class-layer, layer, and class importance heatmaps to interpret how our layers are classifying inputs.

\subsection{Large Movie Review Experiment}

We compare the MLP network with a NFT-K MLP network with two different depths shown in Table~\ref{tab:results}. From Table~\ref{tab:results}, the NFT-K MLP networks perform even more closely to the MLP networks, varying by at most 0.42\%. Thus we can use NFT-K MLPs to perform similarly to MLPs while gaining interpretability. 

The sparsity of $\Omega_n$ in this experiment is close to that of MNIST: 50\% and 45\% of the elements were zero for the 5 layer and 10 layer, respectively. What is even more interesting is that the weights for two layers where completely zeroed out and not used in the NFT-K MLPs. This sparsity of the weights and of the columns of $\Omega$ shows how one can achieve almost the same performance without using a significant part of the network, similar to dropout. \footnote{Our code can be found at \url{https://gitlab.com/cjbarb7/icassp2022}.}

\section{Discussion}
\label{sec:discussion}
Inspired from the NTK theory, we designed a new network that allows connections from all the layers.  From the interpretable weights from the NFT-K network classifier we were able to quantify which layers contributed based on Figure~\ref{fig:mnist}. With these weights, they provide insight to see if it is the early layers contributing the most or the later layers contributing the most. There are many avenues to pursue for future work such as investigating how increasing the depth or the number of classes will affect the prioritization of layers for the classifier. Another question that arises is whether we can apply other sparsity constraints to investigate if more sparse solutions exist with fewer layers contributing to the classification.


\vfill\pagebreak

\small
\bibliography{refs}
\bibliographystyle{ieeetr}

\end{document}